\documentclass[]{article}
\usepackage{lmodern}
\usepackage{amssymb,amsmath}
\usepackage{ifxetex,ifluatex}
\usepackage{fixltx2e} 
\usepackage[T1]{fontenc}
\usepackage[utf8]{inputenc}
\usepackage[unicode=true]{hyperref}
\usepackage[usenames,dvipsnames]{color}
\usepackage{graphicx}
\usepackage{subcaption}
\usepackage{cleveref}
\usepackage{booktabs}
\usepackage{multirow}

\hypersetup{breaklinks=true,
            bookmarks=true,
            pdfauthor={Cezary Kaliszyk, Fran\c{c}ois Chollet, Christian Szegedy},
            pdftitle={HolStep: A Machine Learning Dataset for Higher-order Logic Theorem Proving},
            colorlinks=true,
            citecolor=blue,
            urlcolor=blue,
            linkcolor=magenta,
            colorlinks=false,pdfborder={.5 .5 .5}
            }
\usepackage{natbib}
\title{HolStep: A Machine Learning Dataset for Higher-order Logic Theorem Proving}
\author{Cezary Kaliszyk\\
University of Innsbruck\\
\texttt{cezary.kaliszyk@uibk.ac.at} \\
\And
Fran\c{c}ois Chollet, Christian Szegedy\\
Google Research\\
\texttt{\{fchollet,szegedy\}@google.com}
}

\date{}
\usepackage{iclr2017_conference,times}
\iclrfinalcopy

\ifx\paragraph\undefined\else
\let\oldparagraph\paragraph
\renewcommand{\paragraph}[1]{\oldparagraph{#1}\mbox{}}
\fi
\ifx\subparagraph\undefined\else
\let\oldsubparagraph\subparagraph
\renewcommand{\subparagraph}[1]{\oldsubparagraph{#1}\mbox{}}
\fi

\begin{document}
\maketitle
\begin{abstract}
Large computer-understandable proofs consist of millions of intermediate
logical steps. The vast majority of such steps originate from manually
selected and manually guided heuristics applied to intermediate goals.
So far, machine learning has generally not been used to filter or
generate these steps. In this paper, we introduce a new dataset based on
Higher-Order Logic (HOL) proofs, for the purpose of developing new
machine learning-based theorem-proving strategies. We make this dataset
publicly available under the BSD license. We propose various machine
learning tasks that can be performed on this dataset, and discuss their
significance for theorem proving. We also benchmark a set of simple baseline
machine learning models suited for the tasks (including logistic regression,
convolutional neural networks and recurrent neural networks). The results of our
baseline models show the promise of applying machine learning to HOL
theorem proving.
\end{abstract}

\section{Introduction}\label{s:intro}


As the usability of interactive theorem proving (ITP) systems
\citep{itp} grows, its use becomes a more common way of establishing the
correctness of software as well as mathematical proofs. Today, ITPs are used
for software certification projects ranging from compilers
\citep{compcert} and operating system components
\citep{fscq,sel4}, to establishing the absolute correctness of
large proofs in mathematics such as the Kepler conjecture
\citep{flyspeckfinal} and the Feit-Thomson Theorem \citep{oddorder}.

For results of such significance to be possible, the theorem libraries
of these ITPs must contain all necessary basic mathematical properties,
accompanied with formal proofs. This means that the size of many ITP
libraries can be measured in dozens of thousands of theorems
\citep{mizarnut,afp} and billions of individual proof steps.
While the general direction of the proofs is specified by humans (by
providing the goal to prove, specifying intermediate steps, or applying
certain automated tactics), the majority of such proof steps are
actually found by automated reasoning-based proof search \citep{lemmas},
with very little application of machine learning techniques so far.

At the same time, fast progress has been unfolding in machine learning applied to tasks
that involve logical inference, such as natural language question answering
\citep{sukhbaatar2015end}, knowledge base completion \citep{socher2013}, automated
translation \citep{google_machinetranslation}, and premise selection in the context of
theorem proving \citep{deepmath}. Deep learning in particular has proven
to be a powerful tool for embedding semantic meaning and logical relationships
into geometric spaces, specifically via models such as convolutional
neural networks, recurrent neural networks, and tree-recursive neural
networks. These advances strongly suggest that deep learning may have
become mature enough to yield significant advances in automated theorem
proving. Remarkably, it has recently become possible to build a system,
AlphaGo \citep{alphago}, blending classical AI techniques such as
Monte-Carlo tree search and modern deep learning techniques, capable of
playing the game of Go at super-human levels. We should note that
theorem proving and Go playing are conceptually related, since both
consist in searching for specific nodes in trees of states with
extremely large arity and relatively large depth, which involves node
evaluation decision (how valuable is this state?) and policy decisions
(which node should be expanded next?). The success of AlphaGo can thus
serve as encouragement on the road to building deep learning-augmented
theorem provers that would blend classical techniques developed over the
past few decades with the latest machine learning advances.

Fast progress in specific machine learning verticals has occasionally
been achieved thanks to the release of specialized datasets (often with
associated competitions, e.g.~the ImageNet dataset for large-scale image
classification \citep{imagenet}) serving as an experimental testbed and
public benchmark of current progress, thus focusing the efforts of the
research community. We hope that releasing a theorem proving dataset
suited for specific machine learning tasks can serve the same purpose in
the vertical of applying machine learning to theorem proving.

\subsection{Contribution and Overview}\label{contribution-and-overview}

First, we develop a dataset for machine learning based on the proof
steps used in a large interactive proof \autoref{s:extraction}. We focus
on the HOL Light \citep{hollight} ITP, its multivariate analysis library
\citep{multivariate}, as well as the formal proof of the Kepler
conjecture \citep{flyspeck}. These formalizations constitute a diverse
proof dataset containing basic mathematics, analysis, trigonometry, as
well as reasoning about data structures such as graphs. Furthermore
these formal proof developments have been used as benchmarks for
automated reasoning techniques \citep{holyhammer}.

The dataset consists of 2,013,046 training examples and 196,030 testing
examples that originate from 11,400 proofs.
Precisely half of the examples are statements that were useful in the
currently proven conjectures and half are steps that have been derived
either manually or as part of the automated proof search but were not
necessary in the final proofs. The dataset contains only
proofs of non-trivial theorems, that also do not focus on computation
but rather on actual theorem proving. For each proof, the conjecture
that is being proven as well as its dependencies (axioms) and may be
exploited in machine learning tasks. Furthermore, for each statement
both its human-readable (pretty-printed) statement and a tokenization
designed to make machine learning tasks more manageable are included.

Next, in \autoref{s:tasks} we discuss the proof step classification
tasks that can be attempted using the dataset, and we discuss the
usefulness of these tasks in interactive and automated theorem proving.
These tasks include unconditioned classification (without access to
conjectures and dependencies) and conjecture-conditioned classification
(with access to the conjecture) of proof steps as being useful or not in
a proof. We outline the use of such classification capabilities for
search space pruning and internal guidance, as well as for generation of
intermediate steps or possible new lemma statements.

Finally, in \autoref{s:models} we propose three baseline models for the
proof step classification tasks, and we experimentally evaluate the models
on the data in \autoref{s:results}. The models considered include both a
relatively simple regression model, as well as deep learning models
based on convolutional and recurrent neural networks.

\subsection{Related Work}\label{s:related}

The use of machine learning in interactive and automated theorem proving
has so far focused on three tasks: premise selection, strategy
selection, and internal guidance. We shortly explain these.

Given a large library of proven facts and a user given conjecture,
the multi-label classification problem of selecting the facts
that are most likely to lead to a successful proof of the conjecture has
been usually called \emph{relevance filtering} or \emph{premise
selection} \citep{premsel}. This is crucial for the efficiency of
modern automation techniques for ITPs \citep{hammers}, which today can
usually solve 40--50\% of the conjectures in theorem proving libraries.
Similarly most competitive ATPs today \citep{casc}
implement the SInE classifier \citep{sine}.

A second theorem proving task where machine learning has been of
importance is \emph{strategy selection}. With the development of
automated theorem provers came many parameters that control their
execution. In fact, modern ATPs, such as E \citep{eprover} and
Vampire \citep{vampire}, include complete strategy description languages
that allow a user to specify the orderings, weighting functions, literal
selection strategies, etc. Rather than optimizing the search strategy
globally, one can choose the strategy based on the currently considered
problem. For this some frameworks use machine learning
\citep{learnheuristic,males}.

Finally, an automated theorem prover may use machine learning for
choosing the actual inference steps.
It has been shown to significantly reduce the proof
search in first-order tableaux by the selection of extension steps to
use \citep{malecop}, and has been also successfully applied in
monomorphic higher-order logic proving \citep{mllax}. Data/proof mining
has also been applied on the level of interactive theorem proving
tactics \citep{hazelduncan} to extract and reuse repeating patterns.


\section{Dataset Extraction}\label{s:extraction}

We focus on the HOL Light theorem prover for
two reasons. First, it follows the LCF approach\footnote{LCF approach is
a software architecture for implementing theorem provers which uses a
strongly typed programming language with abstract datatypes (such as
OCaml in the case of HOL Light) to separate the small trusted core,
called the kernel, which
verifies the primitive inferences from user code which allows the user
to arbitrarily extend the system in a safe manner. For more details
see~\citep{edin-lcf}.}). This means that complicated inferences are
reduced to the most primitive ones and the data extraction related
modifications can be restricted the primitive inferences and it is
relatively easy to extract proof steps at
an arbitrary selected level of granularity. Second, HOL Light implements
higher-order logic~\citep{churchhol} as its foundation,
which on the one hand is powerful
enough to encode most of today's formal proofs, and on the other hand
allows for an easy integration of many powerful automation mechanisms
\citep{baadernipkow,blast}.

When selecting the theorems to record, we choose an intermediate approach
between HOL Light ProofRecording \citep{importold} and the HOL/Import one
\citep{importnew}. The theorems that are derived by most common proof
functions are extracted by patching these functions like in the former approach,
and the remaining theorems are extracted from the underlying OCaml programming
language interpreter. In certain cases decision procedures derive theorems
to be reused in subsequent invocations. We detect such values by looking
at theorems used across proof blocks and avoid extracting such reused unrelated
subproofs.

All kernel-level inferences are recorded together with their respective arguments
in a trace file. The trace is processed offline to extract the dependencies of the facts,
detect used proof boundaries,  mark the used and unused steps, and mark the training
and testing examples. Only proofs that have sufficiently many used and unused
steps are considered useful for the dataset. The annotated proof trace
is processed again by a HOL kernel saving the actual training and testing examples
originating from non-trivial reasoning steps. Training and testing examples are grouped by
proof: for each proof the conjecture (statement that is finally proved), the
dependencies of the theorem are constant, and a list of used and not used
intermediate statements is provided. This means that the conjectures used in the
training and testing sets are normally disjoint.

For each statement, whether it is the conjecture, a proof dependency, or an
intermediate statement, both a fully parenthesised HOL Light human-like printout
is provided, as well as a predefined tokenization. The standard HOL Light printer
uses parentheses and operator priorities to make its notations somewhat similar
to textbook-style mathematics, while at the same time preserving the complete
unambiguity of the order of applications (this is particularly visible for
associative operators). The tokenization that we propose attempts to reduce
the number of parentheses. To do this we compute the maximum number of arguments
that each symbol needs to be applied to, and only mark partial application. This
means that fully applied functions (more than 90\% of the applications) do not
require neither application operators nor parentheses. Top-level universal quantifications
are eliminated, bound variables are represented by their de Bruijn indices (the distance from the corresponding abstraction in the parse tree of the term) and free
variables are renamed canonically. Since the Hindley-Milner type inference~\cite{hindley1969principal} mechanisms
will be sufficient to reconstruct the most-general types of the expressions well enough for
automated-reasoning techniques \citet{alignedcorpora} we erase all type information.
Table~\ref{tab:stats} presents some dataset statistics.
%
%
 The dataset, the description of the used format, the scripts used to generate it and baseline models code are available: \\[1mm]
\centerline{\url{http://cl-informatik.uibk.ac.at/cek/holstep/}}

\begin{table}[htb]
  \centering
\setlength{\tabcolsep}{3mm}
\begin{tabular}{lcccc} \toprule
                    & Train & Test & Positive & Negative \\ \midrule
  Examples          & 2013046 & 196030 & 1104538 & 1104538 \\
  Avg. length       & 503.18 & 440.20 & 535.52 & 459.66 \\
  Avg. tokens       & 87.01 & 80.62 & 95.48 & 77.40 \\
  Conjectures       & 9999 & 1411 & - & - \\
  Avg. dependencies & 29.58 & 22.82 & - & - \\ \bottomrule
\end{tabular}
  \caption{HolStep dataset statistics}
  \label{tab:stats}
\end{table}

\section{Machine Learning Tasks}\label{s:tasks}

\subsection{Tasks description}

This dataset makes possible several tasks well-suited for machine
learning most of which are highly relevant for theorem proving:

\begin{itemize}
\item
  Predicting whether a statement is useful in the proof
  of a given conjecture;
\item
  Predicting the dependencies of a proof statement (premise selection);
\item
  Predicting whether a statement is an important one (human named);
\item
  Predicting which conjecture a particular intermediate statement
  originates from;
\item
  Predicting the name given to a statement;
\item
  Generating intermediate statements useful in the proof of a given
  conjecture; 
\item
  Generating the conjecture the current proof will lead to.
\end{itemize}

In what follows we focus on the first task: classifying proof step
statements as being useful or not in the context of a given proof. This
task may be further specialized into two different tasks:

\begin{itemize}
\item
  Unconditioned classification of proof steps: determining how likely a
  given proof is to be useful for the proof it occurred in, based solely
  on the content of statement (i.e.~by only providing the model with the
  step statement itself, absent any context).
\item
  Conditioned classification of proof steps: determining how likely a
  given proof is to be useful for the proof it occurred in, with
  ``conditioning'' on the conjecture statement that the proof was aiming
  to attain, i.e.~by providing the model with both the step statement
  and the conjecture statement).
\end{itemize}

In the dataset, for every proof we provide the same number of useful and
non-useful steps. As such, the proof step classification problem is a
balanced two-class classification problem, where a random baseline would
yield an accuracy of 0.5.

\subsection{Relevance to interactive and automated theorem proving}

In the interaction with an interactive theorem prover, the
tasks that require most human time are: the search for good
intermediate steps; the search for automation techniques able
to justify the individual steps, and searching theorem
proving libraries for the necessary simpler facts. These three
problems directly correspond to the machine learning tasks
proposed in the previous subsection. Being able to predict
the usefulness of a statement will significantly improve many
automation techniques. The generation of good intermediate lemmas
or intermediate steps can improve level of granularity of the
proof steps. Understanding the correspondence between statements
and their names can allow users to search for statements in the
libraries more efficiently \citep{namespark}. Premise selection
and filtering are already used in many theorem proving systems,
and generation of succeeding steps corresponds to conjecturing
and theory exploration.

\section{Baseline Models}\label{s:models}

\begin{figure}[!ht]
  \caption{Unconditioned classification model architectures.}
  \label{uncond_models}
  \centering
    \includegraphics[width=0.6\textwidth]{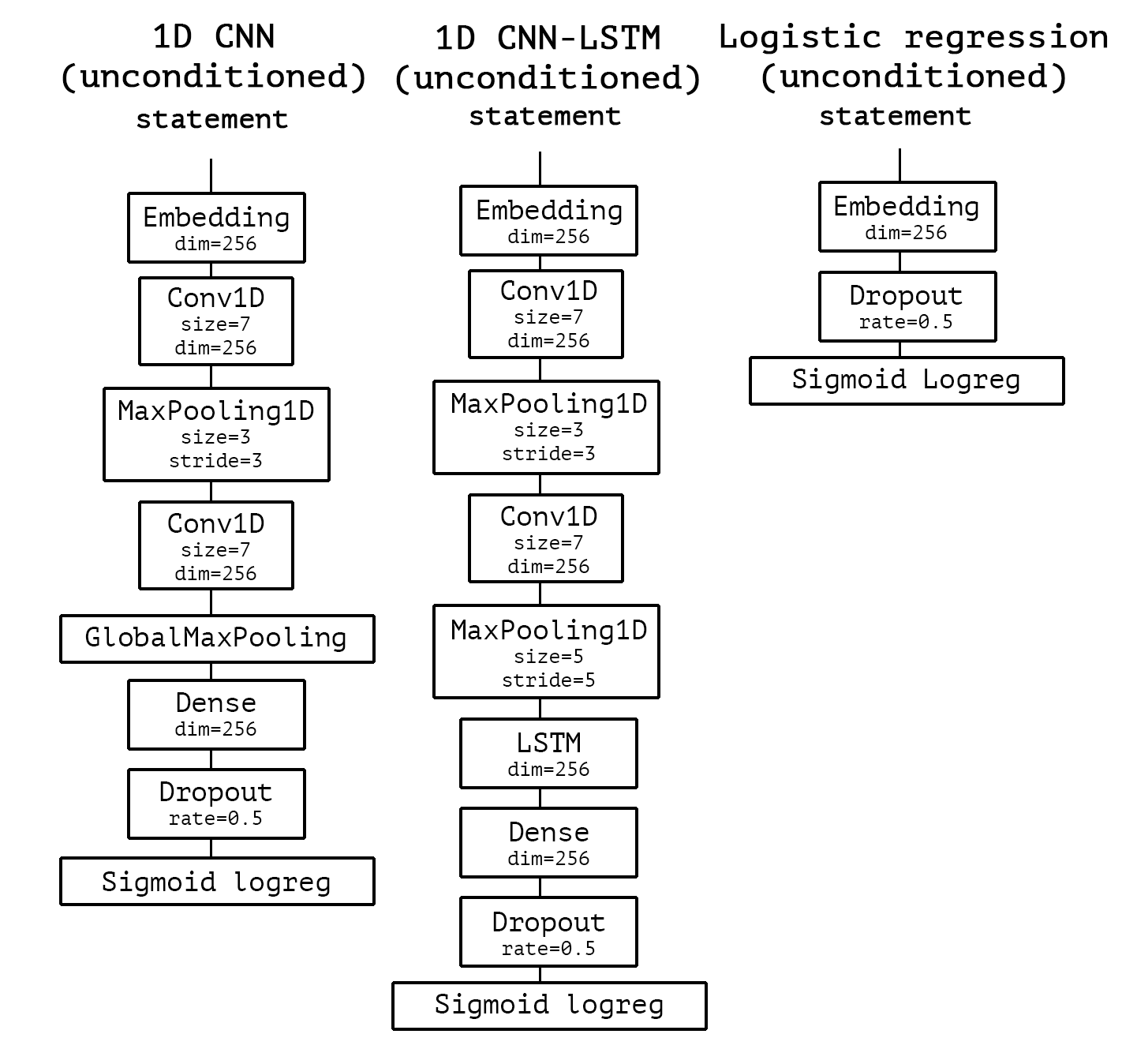}
\end{figure}

\begin{figure}[!ht]
  \caption{Conditioned classification model architectures.}
  \label{cond_models}
  \centering
    \includegraphics[width=0.8\textwidth]{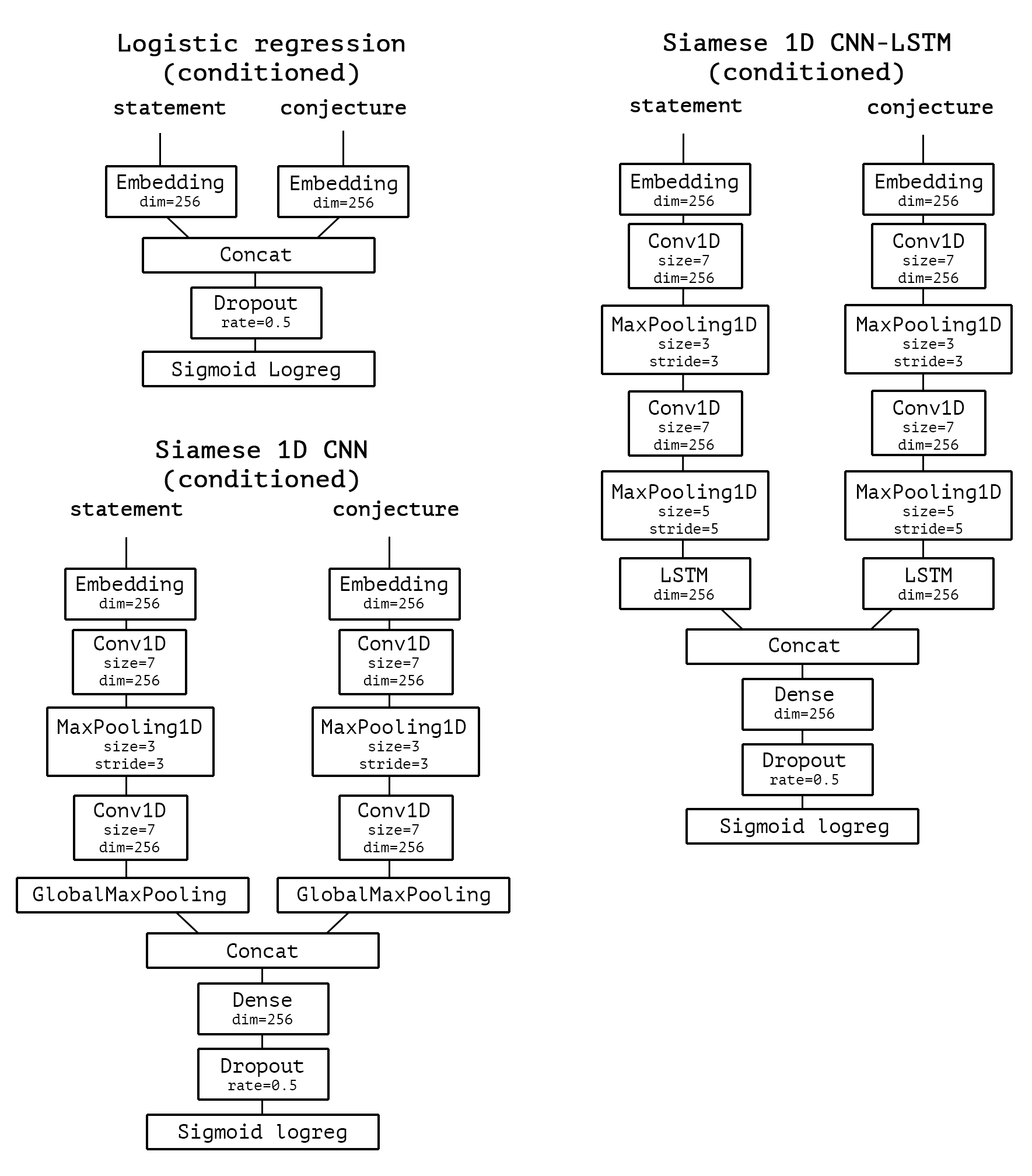}
\end{figure}

For each task (conditioned and unconditioned classification), we propose
three different deep learning architectures, meant to provide a baseline
for the classification performance that can be achieved on this dataset.
Our models cover a range of architecture features (from convolutional
networks to recurrent networks), aiming at probing what characteristics
of the data are the most helpful for usefulness classification.

Our models are implemented in TensorFlow \citep{tensorflow} using the
Keras framework \citep{keras}. Each model was trained on a single Nvidia
K80 GPU. Training only takes a few hours per model, which makes running
these experiments accessible to most people (they could even be run on a
laptop CPU). We are releasing all of our benchmark code as open-source
software
\footnote{\url{https://github.com/tensorflow/deepmath/tree/master/holstep_baselines}}
 so as to allow others to reproduce our results and improve upon
our models.

\subsection{Unconditioned classification models}

Our three models for this task are as follow:

\begin{itemize}
\item
  Logistic regression on top of learned token embeddings. This minimal
  model aims to determine to which extent simple differences between
  token distribution between useful and non-useful statements can be
  used to distinguish them. It provides an absolute floor on the
  performance achievable on this task.
\item
  2-layer 1D convolutional neural network (CNN) with global maxpooling
  for sequence reduction. This model aims to determine the importance of
  local patterns of tokens.
\item
  2-layer 1D CNN with LSTM \citep{lstm} sequence reduction. This model
  aims to determine the importance of order in the features sequences.
\end{itemize}

See figure \ref{uncond_models} for a layer-by-layer description of these models.

\subsection{Conditioned classification models}

For this task, we use versions of the above models that have two siamese
branches (identical branches with shared weights), with one branch
processing the proof step statement being considered, and the other
branch processing the conjecture. Each branch outputs an embedding;
these two embeddings (step embedding and conjecture embedding) are then
concatenated and the classified by a fully-connected network.
See figure \ref{cond_models} for a layer-by-layer description of these models.

\subsection{Input statements encoding}

It should be noted that all of our models start with an Embedding layer,
mapping tokens or characters in the statements to dense vectors in a
low-dimensional space. We consider two possible encodings for presenting
the input statements (proof steps and conjectures) to the Embedding
layers of our models:

\begin{itemize}
\item
  Character-level encoding of the human-readable versions of the
  statements, where each character (out of a set of 86 unique
  characters) in the pretty-printed statements is mapped to a
  256-dimensional dense vector. This encoding yields longer statements
  (training statements are 308 character long on average).
\item
  Token-level encoding of the versions of the statements rendered with
  our proposed high-level tokenization scheme. This encoding yields
  shorter statements (training statements are 60 token long on average),
  while considerably increasing the size of set of unique tokens
  (1993 total tokens in the training set).
\end{itemize}

\section{Results}\label{s:results}

\begin{table}[]
\centering
\caption{HolStep proof step classification accuracy without conditioning}
\label{uncond_results_table}
\setlength{\tabcolsep}{3mm}
  \begin{tabular}{lccc}\toprule
                                    & \textbf{Logistic} & \multirow{2}{*}{\textbf{1D CNN}} &     \multirow{2}{*}{\textbf{1D CNN-LSTM}}   \\
& \textbf{regression} \\ \midrule
  \textbf{Accuracy with char input}     &         0.71            &      0.82             &      \textbf{0.83}         \\
  \textbf{Accuracy with token input}    &         0.71            &       \textbf{0.83}   &            0.77            \\ \bottomrule
  \end{tabular}
\end{table}

\begin{table}[]
\centering
\caption{HolStep proof step classification accuracy with conditioning}
\label{cond_results_table}
\setlength{\tabcolsep}{3mm}
  \begin{tabular}{lccc}
  \toprule
                                    & \textbf{Logistic } & \textbf{Siamese} & \textbf{Siamese}  \\
  & \textbf{regression} & \textbf{1D CNN} & \textbf{1D CNN-LSTM} \\ \midrule
  \textbf{Accuracy with char input}    &          0.71             &       0.81              &      \textbf{0.83}        \\
  \textbf{Accuracy with token input}   &          0.71             &        0.82             &            0.77           \\ \bottomrule
  \end{tabular}
\end{table}

\begin{figure}[!ht]
\centering
  \parbox{.48\textwidth}{
    \hspace{.05\textwidth}\includegraphics[width=.8\linewidth]{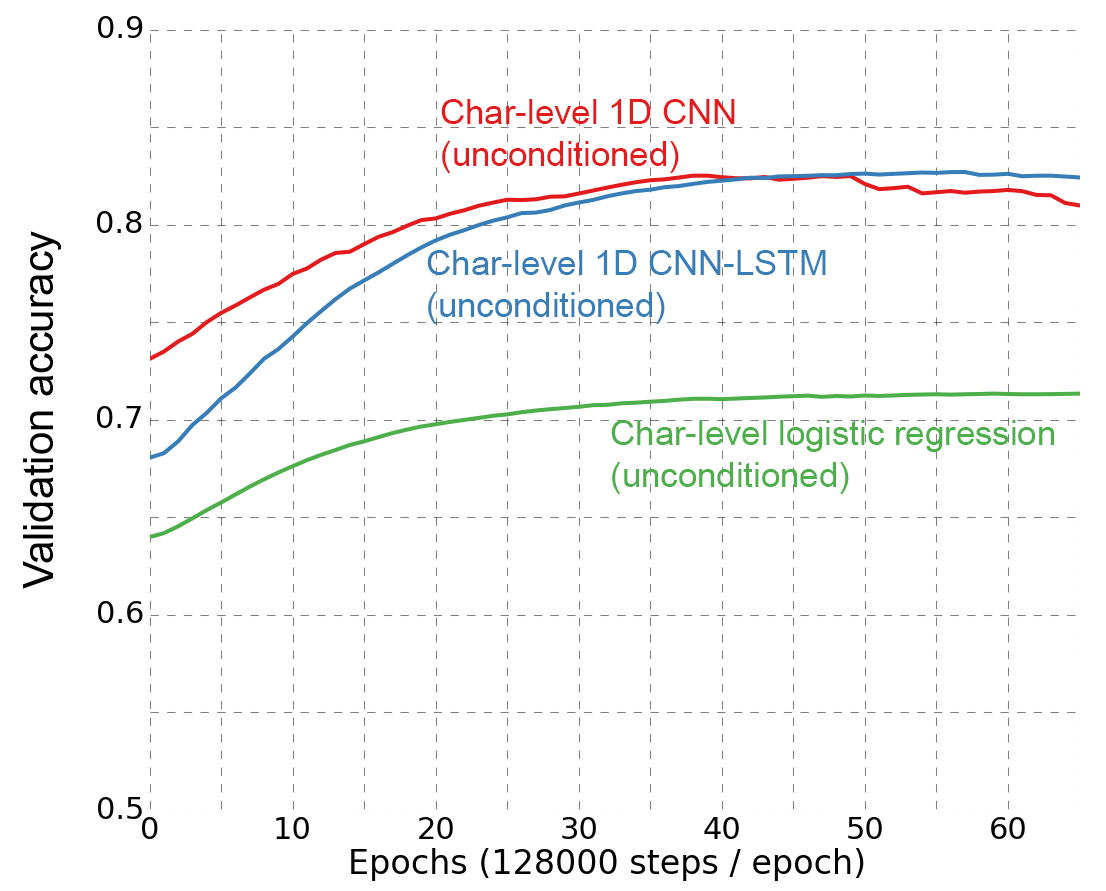}
    \caption{Training profile of the three unconditioned baseline models with character input.}
    \label{uncond_char_fig}
  }\hspace{0.03\textwidth}
  \begin{minipage}{.48\textwidth}
    \hspace{.05\textwidth}\includegraphics[width=.8\linewidth]{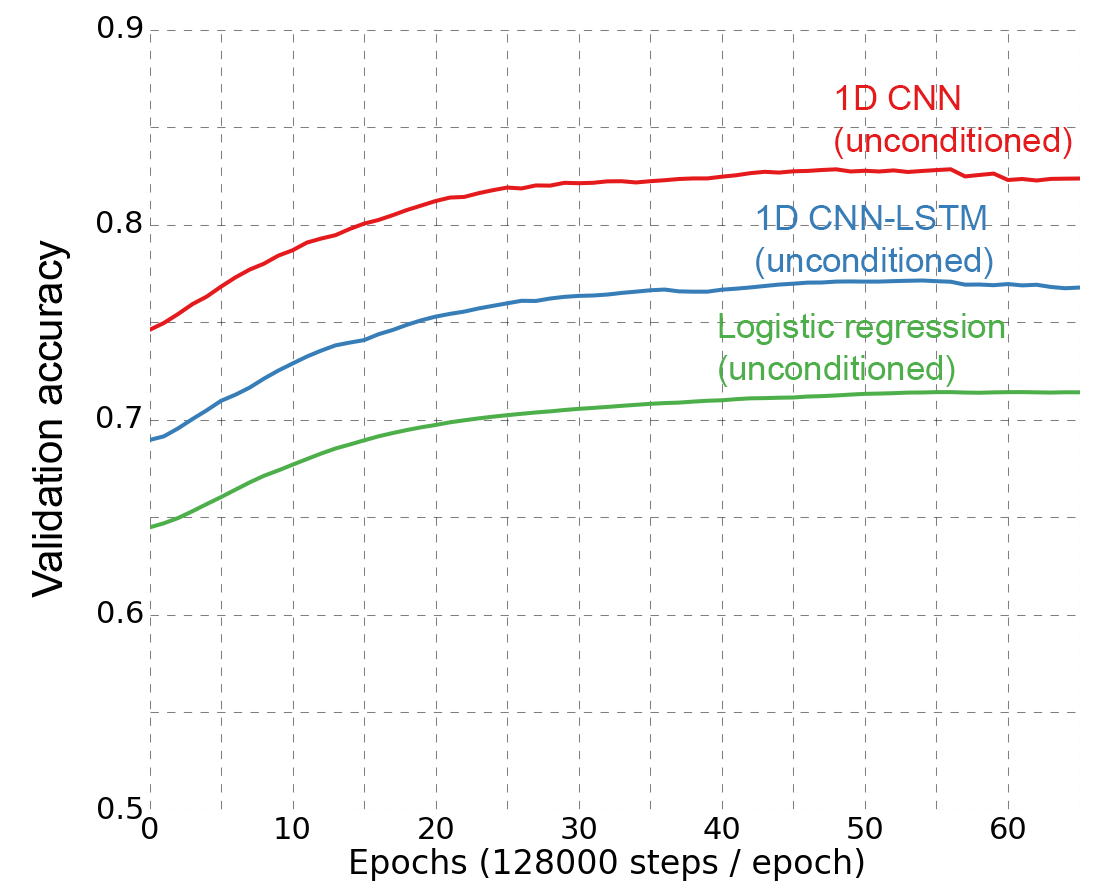}
    \caption{Training profile of the three unconditioned baseline models with token input.}
    \label{uncond_token_fig}
  \end{minipage}
\end{figure}

Experimental results are presented in \cref{uncond_results_table,cond_results_table}, as well as \cref{uncond_char_fig,uncond_token_fig,cond_char_fig,cond_token_fig}.

\subsection{Influence of model architecture}

Our unconditioned logistic regression model yields an accuracy of 71\%,
both with character encoding and token encoding (\cref{uncond_results_table,cond_results_table}).
This demonstrates that differences in token or character distributions between useful and
non-useful steps alone, absent any context, is sufficient for
discriminating between useful and non-useful statements to a reasonable
extent. This also demonstrates that the token encoding is not
fundamentally more informative than raw character-level statements.

Additionally, our unconditioned 1D CNN model yields an accuracy of 82\%
to 83\%, both with character encoding and token encoding (\cref{uncond_results_table,cond_results_table}).
This demonstrates that patterns of characters or patterns of tokens are
considerably more informative than single tokens for the purpose of
usefulness classification.

Finally, our unconditioned convolutional-recurrent model does not
improve upon the results of the 1D CNN, which indicates that our models
are not able to meaningfully leverage order in the feature sequences
into which the statements are encoded.

\subsection{Influence of input encoding}

For the logistic regression model and the 2-layer 1D CNN model, the
choice of input encoding seems to have little impact. For the
convolutional-recurrent model, the use of the high-level tokenization
seems to cause a large decrease in model performance (\cref{uncond_token_fig,cond_token_fig}).
This may be due to the fact that token encoding yields shorter sequences, making the use of
a LSTM less relevant.

\subsection{Influence of conditioning on the conjecture}

\begin{figure}[!ht]
\centering
  \parbox{.48\textwidth}{
    \hspace{.05\textwidth}\includegraphics[width=.8\linewidth]{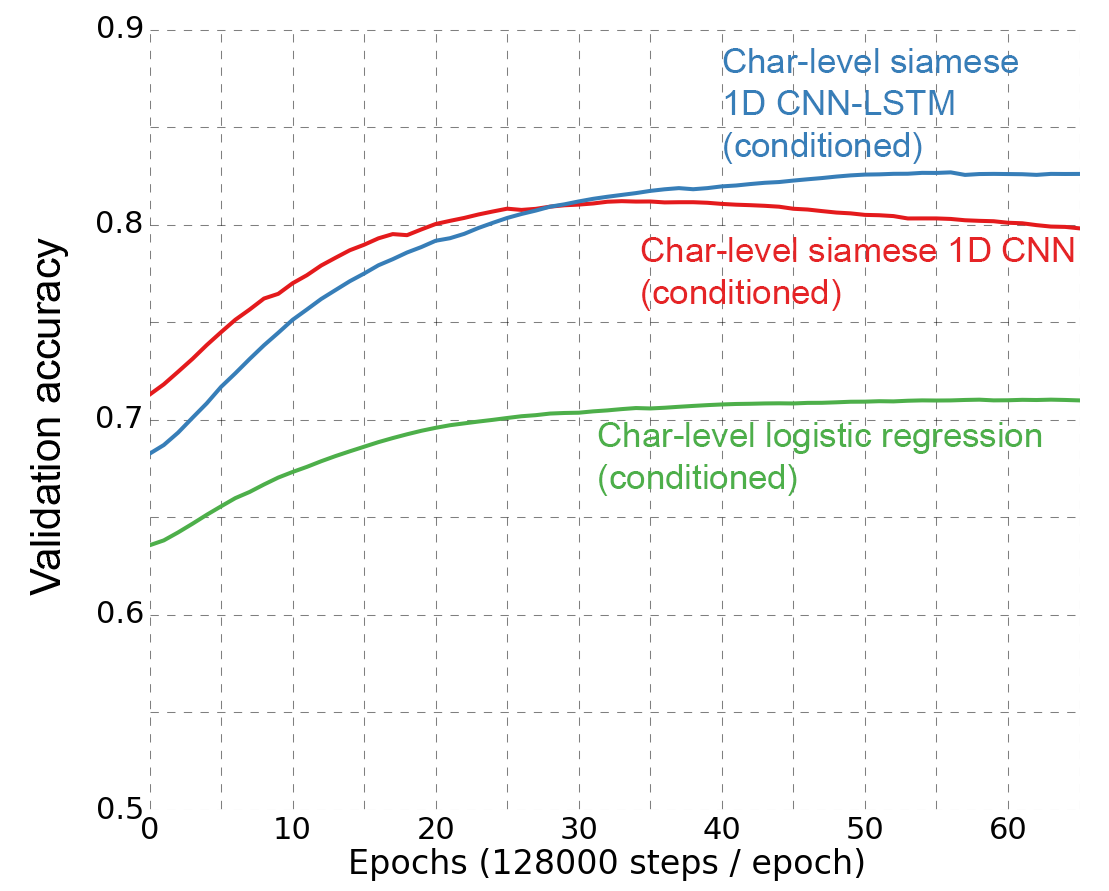}
    \caption{Training profile of the three conditioned baseline models with character input.}
    \label{cond_char_fig}
  }\hspace{0.03\textwidth}
  \begin{minipage}{.48\textwidth}
    \hspace{.05\textwidth}\includegraphics[width=.8\linewidth]{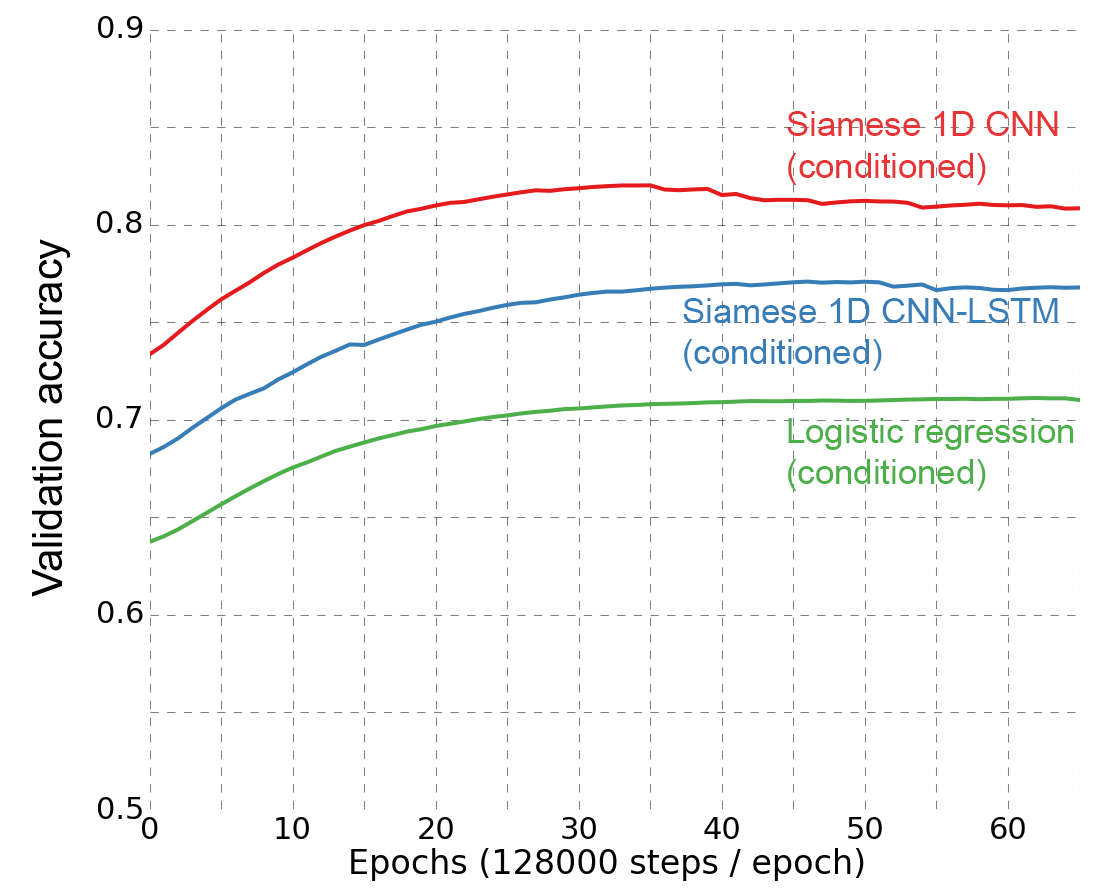}
    \caption{Training profile of the three conditioned baseline models with token input.}
    \label{cond_token_fig}
  \end{minipage}
\end{figure}

None of our conditioned models appear to be able to improve upon the
unconditioned models, which indicates that our architectures are not
able to leverage the information provided by the conjecture. The
presence of the conditioning does however impact the training profile of
our models, in particular by making the 1D CNN model converge faster and
overfit significantly quicker (\cref{cond_char_fig,cond_token_fig}).

\section{Conclusions}\label{s:concl}

Our baseline deep learning models, albeit fairly weak, are still able to
predict statement usefulness with a remarkably high accuracy. Such methods
already help first-order automated provers \citep{femalecop} and as the branching factor
is higher in HOL the predictions are valuable for a number of practical
proving applications.
This includes making tableaux-based \citep{blast} and superposition-based \citep{metis}
internal ITP proof search significantly more efficient in turn making formalization easier.
However, our models do not appear to be able to leverage order in the
input sequences, nor 
 conditioning
on the conjectures. This is due to the fact that these models are not
doing any form of logical reasoning on their input statements; rather
they are doing simple pattern matching at the level of n-grams of
characters or tokens.
This shows the need to focus future efforts on different
models that can do \emph{reasoning}, or alternatively, on systems that
blend explicit reasoning (e.g. graph search) with deep learning-based feature learning.
A potential new direction would be to leverage the graph structure of HOL statements
using e.g. Recursive Neural Tensor Networks \citep{socher2013, recursivenets}
or other graph-based recursive architectures.

\subsection{Future work}\label{future}



The dataset focuses on one interactive theorem prover. It would be interesting
if the proposed techniques generalize, primarily across ITPs that use the same
foundational logic, for example using OpenTheory \citep{opentheory}, and secondarily
across fundamentally different ITPs or even ATPs. A significant part of the unused steps
originates from trying to fulfill the conditions for rewriting and
from calls to intuitionistic tableaux.
The main focus is however on the human found proofs so the trained predictions may
to an extent mimic the bias on the usefulness in the human proofs. As ATPs are at the
moment very week in comparison with human intuition improving this even for the many
proofs humans do not find difficult would be an important gain.

Finally, two of the proposed task for the dataset have been premise selection
and intermediate sentence generation. It would be interesting to
define more ATP-based ways to evaluate the selected premises, as well as to evaluate
generated sentences \citep{alignedcorpora}. The set is a relatively large one when
it comes to proof step classification,
however the number of available premises makes the set a
medium-sized set for premise selection in comparison with those of the  Mizar
Mathematical Library or the seL4 development.

\subsection*{Acknowledgements}
The first author was partly supported by the ERC starting grant 714034.

\bibliography{biblio}

\begin{thebibliography}{46}
\providecommand{\natexlab}[1]{#1}
\providecommand{\url}[1]{\texttt{#1}}
\expandafter\ifx\csname urlstyle\endcsname\relax
  \providecommand{\doi}[1]{doi: #1}\else
  \providecommand{\doi}{doi: \begingroup \urlstyle{rm}\Url}\fi

\bibitem[Abadi et~al.(2015)Abadi, Agarwal, Barham, Brevdo, Chen, Citro,
  Corrado, Davis, Dean, Devin, Ghemawat, Goodfellow, Harp, Irving, Isard, Jia,
  Jozefowicz, Kaiser, Kudlur, Levenberg, Man\'{e}, Monga, Moore, Murray, Olah,
  Schuster, Shlens, Steiner, Sutskever, Talwar, Tucker, Vanhoucke, Vasudevan,
  Vi\'{e}gas, Vinyals, Warden, Wattenberg, Wicke, Yu, and Zheng]{tensorflow}
Mart\'{\i}n Abadi, Ashish Agarwal, Paul Barham, Eugene Brevdo, Zhifeng Chen,
  Craig Citro, Greg~S. Corrado, Andy Davis, Jeffrey Dean, Matthieu Devin,
  Sanjay Ghemawat, Ian Goodfellow, Andrew Harp, Geoffrey Irving, Michael Isard,
  Yangqing Jia, Rafal Jozefowicz, Lukasz Kaiser, Manjunath Kudlur, Josh
  Levenberg, Dan Man\'{e}, Rajat Monga, Sherry Moore, Derek Murray, Chris Olah,
  Mike Schuster, Jonathon Shlens, Benoit Steiner, Ilya Sutskever, Kunal Talwar,
  Paul Tucker, Vincent Vanhoucke, Vijay Vasudevan, Fernanda Vi\'{e}gas, Oriol
  Vinyals, Pete Warden, Martin Wattenberg, Martin Wicke, Yuan Yu, and Xiaoqiang
  Zheng.
\newblock {TensorFlow}: Large-scale machine learning on heterogeneous systems,
  2015.
\newblock URL \url{http://tensorflow.org/}.
\newblock Software available from tensorflow.org.

\bibitem[Alama et~al.(2014)Alama, Heskes, K{\"{u}}hlwein, Tsivtsivadze, and
  Urban]{premsel}
Jesse Alama, Tom Heskes, Daniel K{\"{u}}hlwein, Evgeni Tsivtsivadze, and Josef
  Urban.
\newblock Premise selection for mathematics by corpus analysis and kernel
  methods.
\newblock \emph{J. Autom. Reasoning}, 52\penalty0 (2):\penalty0 191--213, 2014.
\newblock \doi{10.1007/s10817-013-9286-5}.

\bibitem[Alemi et~al.(2016)Alemi, Chollet, Irving, Szegedy, and
  Urban]{deepmath}
Alex~A. Alemi, Fran\c{c}ois Chollet, Geoffrey Irving, Christian Szegedy, and
  Josef Urban.
\newblock {DeepMath} -- {D}eep sequence models for premise selection.
\newblock In Daniel~D. Lee, Masashi Sugiyama, Ulrike~V. Luxburg, Isabelle
  Guyon, and Roman Garnett (eds.), \emph{Advances in Neural Information
  Processing Systems (NIPS 2016)}, pp.\  2235--2243, 2016.
\newblock URL \url{https://arxiv.org/abs/1606.04442}.

\bibitem[Aspinall \& Kaliszyk(2016)Aspinall and Kaliszyk]{namespark}
David Aspinall and Cezary Kaliszyk.
\newblock What's in a theorem name?
\newblock In Jasmin~Christian Blanchette and Stephan Merz (eds.),
  \emph{Interactive Theorem Proving (ITP 2016)}, volume 9807 of \emph{LNCS},
  pp.\  459--465. Springer, 2016.
\newblock \doi{10.1007/978-3-319-43144-4}.

\bibitem[Baader \& Nipkow(1998)Baader and Nipkow]{baadernipkow}
Franz Baader and Tobias Nipkow.
\newblock \emph{Term rewriting and all that}.
\newblock Cambridge University Press, 1998.
\newblock ISBN 978-0-521-45520-6.

\bibitem[Blanchette et~al.(2016)Blanchette, Kaliszyk, Paulson, and
  Urban]{hammers}
Jasmin~C. Blanchette, Cezary Kaliszyk, Lawrence~C. Paulson, and Josef Urban.
\newblock Hammering towards {QED}.
\newblock \emph{J. Formalized Reasoning}, 9\penalty0 (1):\penalty0 101--148,
  2016.
\newblock ISSN 1972-5787.
\newblock \doi{10.6092/issn.1972-5787/4593}.

\bibitem[Blanchette et~al.(2015)Blanchette, Haslbeck, Matichuk, and
  Nipkow]{afp}
Jasmin~Christian Blanchette, Maximilian P.~L. Haslbeck, Daniel Matichuk, and
  Tobias Nipkow.
\newblock Mining the {A}rchive of {F}ormal {P}roofs.
\newblock In Manfred Kerber, Jacques Carette, Cezary Kaliszyk, Florian Rabe,
  and Volker Sorge (eds.), \emph{Intelligent Computer Mathematics (CICM 2015)},
  volume 9150 of \emph{LNCS}, pp.\  3--17. Springer, 2015.

\bibitem[Bridge et~al.(2014)Bridge, Holden, and Paulson]{learnheuristic}
James~P. Bridge, Sean~B. Holden, and Lawrence~C. Paulson.
\newblock Machine learning for first-order theorem proving - learning to select
  a good heuristic.
\newblock \emph{J. Autom. Reasoning}, 53\penalty0 (2):\penalty0 141--172, 2014.
\newblock \doi{10.1007/s10817-014-9301-5}.

\bibitem[Chen et~al.(2016)Chen, Ziegler, Chajed, Chlipala, Kaashoek, and
  Zeldovich]{fscq}
Haogang Chen, Daniel Ziegler, Tej Chajed, Adam Chlipala, M.~Frans Kaashoek, and
  Nickolai Zeldovich.
\newblock Using crash {H}oare logic for certifying the {FSCQ} file system.
\newblock In Ajay Gulati and Hakim Weatherspoon (eds.), \emph{{USENIX} 2016}.
  {USENIX} Association, 2016.

\bibitem[Chollet(2015)]{keras}
Fran\c{c}ois Chollet.
\newblock Keras.
\newblock \url{https://github.com/fchollet/keras}, 2015.

\bibitem[Church(1940)]{churchhol}
Alonzo Church.
\newblock A formulation of the simple theory of types.
\newblock \emph{J. Symb. Log.}, 5\penalty0 (2):\penalty0 56--68, 1940.
\newblock \doi{10.2307/2266170}.
\newblock URL \url{http://dx.doi.org/10.2307/2266170}.

\bibitem[Deng et~al.(2009)Deng, Dong, Socher, Li, Li, and Fei-Fei]{imagenet}
J.~Deng, W.~Dong, R.~Socher, L.-J. Li, K.~Li, and L.~Fei-Fei.
\newblock {ImageNet: A Large-Scale Hierarchical Image Database}.
\newblock In \emph{CVPR09}, 2009.

\bibitem[Duncan(2007)]{hazelduncan}
Hazel Duncan.
\newblock \emph{The Use of Data-Mining for the Automatic Formation of Tactics}.
\newblock PhD thesis, University of Edinburgh, 2007.

\bibitem[F{\"{a}}rber \& Brown(2016)F{\"{a}}rber and Brown]{mllax}
Michael F{\"{a}}rber and Chad~E. Brown.
\newblock Internal guidance for {S}atallax.
\newblock In Nicola Olivetti and Ashish Tiwari (eds.), \emph{International
  Joint Conference on Automated Reasoning (IJCAR 2016)}, volume 9706 of
  \emph{LNCS}, pp.\  349--361. Springer, 2016.
\newblock \doi{10.1007/978-3-319-40229-1}.

\bibitem[Gonthier et~al.(2013)Gonthier, Asperti, Avigad, Bertot, Cohen,
  Garillot, Roux, Mahboubi, O'Connor, Biha, Pasca, Rideau, Solovyev, Tassi, and
  Th{\'e}ry]{oddorder}
Georges Gonthier, Andrea Asperti, Jeremy Avigad, Yves Bertot, Cyril Cohen,
  Fran\c{c}ois Garillot, St{\'e}phane~Le Roux, Assia Mahboubi, Russell
  O'Connor, Sidi~Ould Biha, Ioana Pasca, Laurence Rideau, Alexey Solovyev,
  Enrico Tassi, and Laurent Th{\'e}ry.
\newblock A machine-checked proof of the odd order theorem.
\newblock In Sandrine Blazy, Christine Paulin-Mohring, and David Pichardie
  (eds.), \emph{Interactive Theorem Proving (ITP 2013)}, volume 7998 of
  \emph{LNCS}, pp.\  163--179. Springer, 2013.

\bibitem[Gordon et~al.(1979)Gordon, Milner, and Wadsworth]{edin-lcf}
Michael J.~C. Gordon, Robin Milner, and Christopher~P. Wadsworth.
\newblock \emph{Edinburgh {LCF}}, volume~78 of \emph{Lecture Notes in Computer
  Science}.
\newblock Springer, 1979.
\newblock ISBN 3-540-09724-4.
\newblock \doi{10.1007/3-540-09724-4}.
\newblock URL \url{http://dx.doi.org/10.1007/3-540-09724-4}.

\bibitem[Grabowski et~al.(2010)Grabowski, Kornilowicz, and Naumowicz]{mizarnut}
Adam Grabowski, Artur Kornilowicz, and Adam Naumowicz.
\newblock Mizar in a nutshell.
\newblock \emph{J. Formalized Reasoning}, 3\penalty0 (2):\penalty0 153--245,
  2010.
\newblock \doi{10.6092/issn.1972-5787/1980}.

\bibitem[Hales et~al.(2010)Hales, Harrison, McLaughlin, Nipkow, Obua, and
  Zumkeller]{flyspeck}
Thomas Hales, John Harrison, Sean McLaughlin, Tobias Nipkow, Steven Obua, and
  Roland Zumkeller.
\newblock A revision of the proof of the {Kepler Conjecture}.
\newblock \emph{Discrete {\&} Computational Geometry}, 44\penalty0
  (1):\penalty0 1--34, 2010.

\bibitem[Hales et~al.(2015)Hales, Adams, Bauer, Dang, Harrison, Hoang,
  Kaliszyk, Magron, McLaughlin, Nguyen, Nguyen, Nipkow, Obua, Pleso, Rute,
  Solovyev, Ta, Tran, Trieu, Urban, Vu, and Zumkeller]{flyspeckfinal}
Thomas~C. Hales, Mark Adams, Gertrud Bauer, Dat~Tat Dang, John Harrison,
  Truong~Le Hoang, Cezary Kaliszyk, Victor Magron, Sean McLaughlin, Thang~Tat
  Nguyen, Truong~Quang Nguyen, Tobias Nipkow, Steven Obua, Joseph Pleso, Jason
  Rute, Alexey Solovyev, An~Hoai~Thi Ta, Trung~Nam Tran, Diep~Thi Trieu, Josef
  Urban, Ky~Khac Vu, and Roland Zumkeller.
\newblock A formal proof of the {K}epler conjecture.
\newblock \emph{CoRR}, abs/1501.02155, 2015.

\bibitem[Harrison(2009)]{hollight}
John Harrison.
\newblock {HOL Light}: An overview.
\newblock In Stefan Berghofer, Tobias Nipkow, Christian Urban, and Makarius
  Wenzel (eds.), \emph{Theorem Proving in Higher Order Logics (TPHOLs 2009)},
  volume 5674 of \emph{LNCS}, pp.\  60--66. Springer, 2009.

\bibitem[Harrison(2013)]{multivariate}
John Harrison.
\newblock The {HOL} {L}ight theory of {E}uclidean space.
\newblock \emph{J. Autom. Reasoning}, 50\penalty0 (2):\penalty0 173--190, 2013.
\newblock \doi{10.1007/s10817-012-9250-9}.

\bibitem[Harrison et~al.(2014)Harrison, Urban, and Wiedijk]{itp}
John Harrison, Josef Urban, and Freek Wiedijk.
\newblock History of interactive theorem proving.
\newblock In J\"org Siekmann (ed.), \emph{Handbook of the History of Logic vol.
  9 (Computational Logic)}, pp.\  135--214. Elsevier, 2014.

\bibitem[Hindley(1969)]{hindley1969principal}
R.~Hindley.
\newblock {The principal type-scheme of an object in combinatory logic}.
\newblock \emph{Transactions of the american mathematical society},
  146:\penalty0 29--60, 1969.
\newblock ISSN 0002-9947.

\bibitem[Hochreiter \& Schmidhuber(1997)Hochreiter and Schmidhuber]{lstm}
Sepp Hochreiter and J{\"u}rgen Schmidhuber.
\newblock Long short-term memory.
\newblock \emph{Neural computation}, 9\penalty0 (8):\penalty0 1735--1780, 1997.

\bibitem[Hoder \& Voronkov(2011)Hoder and Voronkov]{sine}
Kry\v{s}tof Hoder and Andrei Voronkov.
\newblock Sine qua non for large theory reasoning.
\newblock In Nikolaj Bj{\o}rner and Viorica Sofronie-Stokkermans (eds.),
  \emph{CADE-23}, volume 6803 of \emph{LNAI}, pp.\  299--314. Springer, 2011.

\bibitem[Hurd(2003)]{metis}
Joe Hurd.
\newblock First-order proof tactics in higher-order logic theorem provers.
\newblock In Myla Archer, Ben~Di Vito, and C{\'{e}}sar Mu{\~{n}}oz (eds.),
  \emph{Design and Application of Strategies/Tactics in Higher Order Logics
  (STRATA 2003)}, number NASA/CP-2003-212448 in {NASA} Technical Reports, pp.\
  56--68, September 2003.
\newblock URL \url{http://www.gilith.com/research/papers}.

\bibitem[Hurd(2011)]{opentheory}
Joe Hurd.
\newblock The {OpenTheory} standard theory library.
\newblock In Mihaela~Gheorghiu Bobaru, Klaus Havelund, Gerard~J. Holzmann, and
  Rajeev Joshi (eds.), \emph{NASA Formal Methods (NFM 2011)}, volume 6617 of
  \emph{LNCS}, pp.\  177--191. Springer, 2011.

\bibitem[Kaliszyk \& Krauss(2013)Kaliszyk and Krauss]{importnew}
Cezary Kaliszyk and Alexander Krauss.
\newblock Scalable {LCF}-style proof translation.
\newblock In Sandrine Blazy, Christine Paulin-Mohring, and David Pichardie
  (eds.), \emph{Interactive Theorem Proving (ITP 2013)}, volume 7998 of
  \emph{LNCS}, pp.\  51--66. Springer, 2013.

\bibitem[Kaliszyk \& Urban(2014)Kaliszyk and Urban]{holyhammer}
Cezary Kaliszyk and Josef Urban.
\newblock Learning-assisted automated reasoning with {F}lyspeck.
\newblock \emph{J. Autom. Reasoning}, 53\penalty0 (2):\penalty0 173--213, 2014.
\newblock \doi{10.1007/s10817-014-9303-3}.

\bibitem[Kaliszyk \& Urban(2015{\natexlab{a}})Kaliszyk and Urban]{femalecop}
Cezary Kaliszyk and Josef Urban.
\newblock {FEMaLeCoP}: Fairly efficient machine learning connection prover.
\newblock In Martin Davis, Ansgar Fehnker, Annabelle McIver, and Andrei
  Voronkov (eds.), \emph{20th International Conference on Logic for
  Programming, Artificial Intelligence, and Reasoning (LPAR 2015)}, volume 9450
  of \emph{LNCS}, pp.\  88--96. Springer, 2015{\natexlab{a}}.
\newblock \doi{10.1007/978-3-662-48899-7}.

\bibitem[Kaliszyk \& Urban(2015{\natexlab{b}})Kaliszyk and Urban]{lemmas}
Cezary Kaliszyk and Josef Urban.
\newblock Learning-assisted theorem proving with millions of lemmas.
\newblock \emph{J.~Symbolic Computation}, 69:\penalty0 109--128,
  2015{\natexlab{b}}.
\newblock \doi{10.1016/j.jsc.2014.09.032}.

\bibitem[Kaliszyk et~al.(2015)Kaliszyk, Urban, and
  Vysko\v{c}il]{alignedcorpora}
Cezary Kaliszyk, Josef Urban, and Ji\v{r}\'{\i} Vysko\v{c}il.
\newblock Learning to parse on aligned corpora.
\newblock In Christian Urban and Xingyuan Zhang (eds.), \emph{Proc. 6h
  Conference on Interactive Theorem Proving (ITP'15)}, volume 9236 of
  \emph{LNCS}, pp.\  227--233. Springer-Verlag, 2015.
\newblock \doi{10.1007/978-3-319-22102-1_15}.

\bibitem[Klein et~al.(2014)Klein, Andronick, Elphinstone, Murray, Sewell,
  Kolanski, and Heiser]{sel4}
Gerwin Klein, June Andronick, Kevin Elphinstone, Toby~C. Murray, Thomas Sewell,
  Rafal Kolanski, and Gernot Heiser.
\newblock Comprehensive formal verification of an {OS} microkernel.
\newblock \emph{ACM Trans. Comput. Syst.}, 32\penalty0 (1):\penalty0 2, 2014.

\bibitem[Kov{\'a}cs \& Voronkov(2013)Kov{\'a}cs and Voronkov]{vampire}
Laura Kov{\'a}cs and Andrei Voronkov.
\newblock First-order theorem proving and {V}ampire.
\newblock In Natasha Sharygina and Helmut Veith (eds.), \emph{Computer-Aided
  Verification (CAV 2013)}, volume 8044 of \emph{LNCS}, pp.\  1--35. Springer,
  2013.

\bibitem[K{\"{u}}hlwein \& Urban(2015)K{\"{u}}hlwein and Urban]{males}
Daniel K{\"{u}}hlwein and Josef Urban.
\newblock {MaLeS}: {A} framework for automatic tuning of automated theorem
  provers.
\newblock \emph{J. Autom. Reasoning}, 55\penalty0 (2):\penalty0 91--116, 2015.
\newblock \doi{10.1007/s10817-015-9329-1}.

\bibitem[Leroy(2009)]{compcert}
Xavier Leroy.
\newblock Formal verification of a realistic compiler.
\newblock \emph{Commun. ACM}, 52\penalty0 (7):\penalty0 107--115, 2009.

\bibitem[Obua \& Skalberg(2006)Obua and Skalberg]{importold}
Steven Obua and Sebastian Skalberg.
\newblock Importing {HOL} into {Isabelle/HOL}.
\newblock In Ulrich Furbach and Natarajan Shankar (eds.), \emph{International
  Joint Conference on Automated Reasoning (IJCAR 2006)}, volume 4130 of
  \emph{LNCS}, pp.\  298--302. Springer, 2006.

\bibitem[Paulson(1999)]{blast}
Lawrence~C. Paulson.
\newblock A generic tableau prover and its integration with {Isabelle}.
\newblock \emph{J. Universal Computer Science}, 5\penalty0 (3):\penalty0
  73--87, 1999.

\bibitem[Schulz(2013)]{eprover}
Stephan Schulz.
\newblock System description: {E} 1.8.
\newblock In Kenneth~L. McMillan, Aart Middeldorp, and Andrei Voronkov (eds.),
  \emph{Logic for Programming, Artificial Intelligence (LPAR 2013)}, volume
  8312 of \emph{LNCS}, pp.\  735--743. Springer, 2013.

\bibitem[Silver et~al.(2016)Silver, Huang, Maddison, Guez, Sifre, van~den
  Driessche, Schrittwieser, Antonoglou, Panneershelvam, Lanctot, Dieleman,
  Grewe, Nham, Kalchbrenner, Sutskever, Lillicrap, Leach, Kavukcuoglu, Graepel,
  and Hassabis]{alphago}
David Silver, Aja Huang, Christopher~J. Maddison, Arthur Guez, Laurent Sifre,
  George van~den Driessche, Julian Schrittwieser, Ioannis Antonoglou, Veda
  Panneershelvam, Marc Lanctot, Sander Dieleman, Dominik Grewe, John Nham, Nal
  Kalchbrenner, Ilya Sutskever, Timothy Lillicrap, Madeleine Leach, Koray
  Kavukcuoglu, Thore Graepel, and Demis Hassabis.
\newblock Mastering the game of go with deep neural networks and tree search.
\newblock \emph{Nature}, 529:\penalty0 484--503, 2016.
\newblock URL
  \url{http://www.nature.com/nature/journal/v529/n7587/full/nature16961.html}.

\bibitem[Socher et~al.(2013{\natexlab{a}})Socher, Chen, Manning, and
  Ng]{socher2013}
Richard Socher, Danqi Chen, Christopher~D. Manning, and Andrew~Y. Ng.
\newblock Reasoning with neural tensor networks for knowledge base completion.
\newblock In \emph{Advances in Neural Information Processing Systems 26: 27th
  Annual Conference on Neural Information Processing Systems 2013.
  Proceedings.}, pp.\  926--934, 2013{\natexlab{a}}.
\newblock URL
  \url{http://papers.nips.cc/paper/5028-reasoning-with-neural-tensor-networks-for-knowledge-base-completion}.

\bibitem[Socher et~al.(2013{\natexlab{b}})Socher, Perelygin, Wu, Chuang,
  Manning, Ng, and Potts]{recursivenets}
Richard Socher, Alex Perelygin, Jean Wu, Jason Chuang, Christopher~D. Manning,
  Andrew~Y. Ng, and Christopher Potts.
\newblock Recursive deep models for semantic compositionality over a sentiment
  treebank.
\newblock In \emph{Proceedings of the 2013 Conference on {E}mpirical {M}ethods
  in {N}atural {L}anguage {P}rocessing}, pp.\  1631--1642, Stroudsburg, PA,
  October 2013{\natexlab{b}}. Association for Computational Linguistics.

\bibitem[Sukhbaatar et~al.(2015)Sukhbaatar, Weston, Fergus,
  et~al.]{sukhbaatar2015end}
Sainbayar Sukhbaatar, Jason Weston, Rob Fergus, et~al.
\newblock End-to-end memory networks.
\newblock In \emph{Advances in Neural Information Processing Systems}, pp.\
  2431--2439, 2015.

\bibitem[Sutcliffe(2016)]{casc}
Geoff Sutcliffe.
\newblock The {CADE} {ATP} system competition - {CASC}.
\newblock \emph{{AI} Magazine}, 37\penalty0 (2):\penalty0 99--101, 2016.

\bibitem[Urban et~al.(2011)Urban, Vyskočil, and \v{S}t\v{e}p{\'a}nek]{malecop}
Josef Urban, Ji\v{r}\'{\i} Vyskočil, and Petr \v{S}t\v{e}p{\'a}nek.
\newblock {MaLeCoP}: Machine learning connection prover.
\newblock In Kai Br{\"u}nnler and George Metcalfe (eds.), \emph{TABLEAUX 2011},
  volume 6793 of \emph{LNCS}. Springer, 2011.

\bibitem[Wu et~al.(2016)Wu, Schuster, Chen, Le, Norouzi, Macherey, Krikun, Cao,
  Gao, Macherey, Klingner, Shah, Johnson, Liu, Kaiser, Gouws, Kato, Kudo,
  Kazawa, Stevens, Kurian, Patil, Wang, Young, Smith, Riesa, Rudnick, Vinyals,
  Corrado, Hughes, and Dean]{google_machinetranslation}
Yonghui Wu, Mike Schuster, Zhifeng Chen, Quoc~V. Le, Mohammad Norouzi, Wolfgang
  Macherey, Maxim Krikun, Yuan Cao, Qin Gao, Klaus Macherey, Jeff Klingner,
  Apurva Shah, Melvin Johnson, Xiaobing Liu, Lukasz Kaiser, Stephan Gouws,
  Yoshikiyo Kato, Taku Kudo, Hideto Kazawa, Keith Stevens, George Kurian,
  Nishant Patil, Wei Wang, Cliff Young, Jason Smith, Jason Riesa, Alex Rudnick,
  Oriol Vinyals, Greg Corrado, Macduff Hughes, and Jeffrey Dean.
\newblock Google's neural machine translation system: Bridging the gap between
  human and machine translation.
\newblock \emph{CoRR}, abs/1609.08144, 2016.
\newblock URL \url{http://arxiv.org/abs/1609.08144}.

\end{thebibliography}

\end{document}